\documentclass[a4paper]{article}
\usepackage{pdfpages}
\pdfoutput=1
\usepackage{arxiv}
\usepackage[utf8]{inputenc} 
\usepackage[T1]{fontenc}    
\usepackage{booktabs}       
\usepackage{amsfonts}       
\usepackage{nicefrac}       
\usepackage{microtype}      
\usepackage{lipsum}
\usepackage{authblk}
\usepackage{caption}
\usepackage{graphicx}
\usepackage{url}
\usepackage{parskip}
\usepackage{amsmath,bm}
\usepackage{multirow}
\usepackage[colorlinks = true,
            linkcolor = purple,
            urlcolor  = purple,
            citecolor = blue,
            anchorcolor = black]{hyperref}
\usepackage[english]{babel}
\usepackage[square,sort,comma,numbers]{natbib}

\title{Word Embedding based New Corpus for Low-resourced Language: Sindhi}
\author{Wazir Ali, Jay Kumar, Junyu Lu, Zenglin Xu \\ School of Computer Science and Engineering\\University of Electronic Science and Technology of China}
\begin{document}
\maketitle
\begin{abstract} Representing words and phrases into dense vectors of real numbers which encode semantic and syntactic properties is a vital constituent in natural language processing (NLP). The success of neural network (NN) models in NLP largely rely on such dense word representations learned on the large unlabeled corpus.  Sindhi is one of the rich morphological language, spoken by large population in Pakistan and India lacks corpora which plays an essential role of a test-bed for generating word embeddings and developing language independent NLP systems. In this paper, a large corpus of more than 61 million words is developed for low-resourced Sindhi language for training neural word embeddings. The corpus is acquired from multiple web-resources using web-scrappy. Due to the unavailability of open source preprocessing tools for Sindhi, the prepossessing of such large corpus becomes a challenging problem specially cleaning of noisy data extracted from web resources. Therefore, a preprocessing pipeline is employed for the filtration of noisy text. Afterwards, the cleaned vocabulary is utilized for training Sindhi word embeddings with state-of-the-art GloVe, Skip-Gram (SG), and Continuous Bag of Words (CBoW) word2vec algorithms. The intrinsic evaluation approach of cosine similarity matrix and WordSim-353 are employed for the evaluation of generated Sindhi word embeddings. Moreover, we compare the proposed word embeddings with recently revealed Sindhi fastText (SdfastText) word representations. Our intrinsic evaluation results demonstrate the high quality of our generated Sindhi word embeddings using SG, CBoW, and GloVe as compare to SdfastText word representations.
\end{abstract}

\keywords{Corpus acquisition \and Sindhi language \and Neural networks\and Word embeddings \and Continuous Bag of Words\and Skip gram\and GloVe}

\section{Introduction}
\label{intro}
Sindhi is a rich morphological, mutltiscript, and multidilectal language. It belongs to the Indo-Aryan language family~\cite{cole2006sindhi}, with significant cultural and historical background. Presently, it is recognized as is an official language ~\cite{motlani2016developing} in Sindh province of Pakistan, also being taught as a compulsory subject in Schools and colleges.  Sindhi is also recognized as one of the national languages in India. Ulhasnagar, Rajasthan, Gujarat, and Maharashtra are the largest Indian regions of Sindhi native speakers. It is also spoken in other countries except for Pakistan and India, where native Sindhi speakers have migrated, such as America, Canada, Hong Kong, British, Singapore, Tanzania, Philippines, Kenya, Uganda, and South, and East Africa. Sindhi has rich morphological structure~\cite{shaikh2017instant} due to a large number of homogeneous words. Historically, it was written in multiple writing systems, which differ from each other in terms of orthography and morphology. The Persian-Arabic is the standard script of Sindhi, which was officially accepted in 1852 by the British government\footnote{\url{https://www.britannica.com/topic/Sindhi-language}}. However, the Sindhi-Devanagari is also a popular writing system in India being written in left to right direction like the Hindi language. Formerly, Khudabadi, Gujrati, Landa,  Khojki, and Gurumukhi were also adopted as its writing systems. Even though, Sindhi has great historical and literal background, presently spoken by nearly 75 million people~\cite{motlani2016developing}. The research on SNLP was coined in 2002\footnote{"Sindhia lai Kampyutar jo Istemalu" (Use of computer for Sindhi), an article published in Sindhu yearly, Ulhasnagar. 2002}, however, IT grabbed research attention after the development of its Unicode system~\cite{bhurgri2006enabling}. But still, Sindhi stands among the low-resourced languages due to the scarcity of core language processing resources of the raw and annotated corpus, which can be utilized for training robust word embeddings or the use of machine learning algorithms. Since the development of annotated datasets requires time and human resources. 
\par \setlength{\parindent}{3ex}
The Language Resources (LRs) are fundamental elements for the development of high quality NLP systems based on automatic or NN based approaches. The LRs include written or spoken corpora, lexicons, and annotated corpora for specific computational purposes. The development of such resources has received great research interest for the digitization of human languages~\cite{jamro2017sindhi}. Many world languages are rich in such language processing resources integrated in their software tools including English~\cite{loper2002nltk}~\cite{manning2014stanford}, Chinese~\cite{che2010ltp} and other languages~\cite{popel2010tectomt} ~\cite{padro2010freeling}. The Sindhi language lacks the basic computational resources~\cite{Narejo2016Morphology} of a large text corpus, which can be utilized for training robust word embeddings and developing language independent NLP applications including semantic analysis, sentiment analysis, parts of the speech tagging, named entity recognition, machine translation~\cite{li2018word}, multitasking~\cite{collobert2008unified},~\cite{devlin2019bert}. Presently Sindhi Persian-Arabic is frequently used for online communication, newspapers, public institutions in Pakistan, and India ~\cite{motlani2016developing}. But little work has been carried out for the development of LRs such as raw corpus~\cite{rahman2010towards},~\cite{khoso2019build}, annotated corpus~\cite{dootio2018unicode},~\cite{dootio2019development}, ~\cite{motlani2016developing}, ~\cite{ali2017sentiment}. In the best of our knowledge, Sindhi lacks the large unlabelled corpus which can be utilized for generating and evaluating word embeddings for Statistical Sindhi Language Processing (SSLP)  
\par \setlength{\parindent}{3ex} 
One way to to break out this loop is to learn word embeddings from unlabelled corpora, which can be utilized to bootstrap other downstream NLP tasks. The word embedding is a new term of semantic vector space~\cite{lund1996producing}, distributed representations~\cite{mikolov2013distributed},  and distributed semantic models. It is a language modeling approach~\cite{bengio2003neural} used for the mapping of words and phrases into $n$-dimensional dense vectors of real numbers that effectively capture the semantic and syntactic relationship with neighboring words in a geometric way~\cite{andreas2014much}~\cite{schnabel2015evaluation}. Such as  “Einstein” and “Scientist” would have greater similarity compared with “Einstein” and “doctor.” In this way, word embeddings accomplish the important linguistic concept of ``a word is characterized by the company it keeps". More recently NN based models yield state-of-the-art performance in multiple NLP tasks ~\cite{mikolov2018advances}~~\cite{grave2018learning} with the word embeddings. One of the advantages of such techniques is they use unsupervised approaches for learning representations and do not require annotated corpus which is rare for low-resourced Sindhi language. Such representions can be trained on large unannotated corpora, and then generated representations can be used in the NLP tasks which uses a small amount of labelled data.     
\par \setlength{\parindent}{3ex} 
In this paper, we address the problems of corpus construction by collecting a large corpus of more than 61 million words from multiple web resources using the web-scrappy framework. After the collection of the corpus, we carefully preprocessed for the filtration of noisy text, e.g., the HTML tags and vocabulary of the English language. The statistical analysis is also presented for the letter, word frequencies and identification of stop-words. Finally, the corpus is utilized to generate Sindhi word embeddings using state-of-the-art GloVe~\cite{pennington2014glove} SG and CBoW~\cite{mikolov2013efficient}~\cite{mikolov2013distributed}~\cite{mikolov2018advances} algorithms. The popular intrinsic evaluation method \cite{mikolov2013distributed}~\cite{nayak2016evaluating}~\cite{pierrejean2018towards} of calculating cosine similarity between word vectors and WordSim353~\cite{agirre2009study} are employed to measure the performance of the learned Sindhi word embeddings. We translated English WordSim353\footnote{Available online at https://rdrr.io/cran/wordspace/man/WordSim353.html} word pairs into Sindhi using bilingual English to Sindhi dictionary. The intrinsic approach typically involves a pre-selected set of query terms~\cite{schnabel2015evaluation} and semantically related target words, which we refer to as query words. Furthermore, we also compare the proposed word embeddings with recently revealed Sindhi fastText  (SdfastText)\footnote{We denote Sindhi word representations as (SdfastText) recently revealed by fastText, available at (https://fasttext.cc/docs/en/crawl-vectors.html) trained on Common Crawl and Wikipedia corpus of Sindhi Persian-Arabic.}~\cite{grave2018learning} word representations. To the best of our knowledge, this is the first comprehensive work on the development of large corpus and generating word embeddings along with systematic evaluation for low-resourced Sindhi Persian-Arabic. The synopsis of our novel contributions is listed as follows:
\begin{itemize}
   \item We present a large corpus of more than 61 million words obtained from multiple web resources and reveal a list of Sindhi stop words. 
    \item We develop a text cleaning pipeline for the preprocessing of the raw corpus.  
    \item  Generate word embeddings using GloVe, CBoW, and SG Word2Vec algorithms also evaluate and compare them using the intrinsic evaluation approaches of cosine similarity matrix and WordSim353.
    \item  We are the first to evaluate SdfastText word representations and compare them with our proposed Sindhi word embeddings. 
\end{itemize}
The remaining sections of the paper are organized as, Section \ref{relatedwork} presents the literature survey regarding computational resources, Sindhi corpus construction, and word embedding models. Afterwards, Section \ref{sec:methodology} presents the employed methodology, Section \ref{sec:statistical_analysis_of_corpus} consist of statistical analysis of the developed corpus. Section \ref{sec:exp_set} present the experimental setup. The intrinsic evaluation results along with comparison are given in Section \ref{sec:word_similarity_comparision}. The discussion and future work are given in Section \ref{sec:discussion}, and lastly, Section \ref{sec:conclusion} presents the conclusion.
\section{Related work}
\label{relatedwork}
The natural language resources refer to a set of language data and descriptions~\cite{schafer2013web} in machine readable form, used for building, improving, and evaluating NLP algorithms or softwares.  Such resources include written or spoken corpora, lexicons, and annotated corpora for specific computational purposes. Many world languages are rich in such language processing resources integrated in the software tools including  NLTK for  English~\cite{loper2002nltk}, Stanford CoreNLP~\cite{manning2014stanford}, LTP for Chinese~\cite{che2010ltp}, TectoMT for German, Russian, Arabic~\cite{popel2010tectomt} and multilingual toolkit~\cite{padro2010freeling}. But Sindhi language is at an early stage for the development of such resources and software tools. 
\par \setlength{\parindent}{3ex} 
The corpus construction for NLP mainly involves important steps of acquisition, preprocessing, and tokenization. Initially,~\cite{rahman2010towards} discussed the morphological structure and challenges concerned with the corpus development along with orthographical and morphological features in the Persian-Arabic script. The raw and annotated corpus~\cite{motlani2016developing} for Sindhi Persian-Arabic is a good supplement towards the development of resources, including raw and annotated datasets for parts of speech tagging, morphological analysis, transliteration between Sindhi Persian-Arabic and Sindhi-Devanagari, and machine translation system. But the corpus is acquired only form Wikipedia-dumps. A survey-based study~\cite{jamro2017sindhi} provides all the progress made in the Sindhi Natural Language Processing (SNLP) with the complete gist of adopted techniques, developed tools and available resources which show that work on resource development on Sindhi needs more sophisticated efforts. \textbf{The raw corpus is utilized for Sindhi word segmentation~\cite{bhatti2014word}.} More recently, an initiative towards the development of resources is taken~\cite{dootio2018unicode} by open sourcing annotated dataset of Sindhi Persian-Arabic obtained from news and social blogs. The existing and proposed work is presented in Table \ref{tab:related_work} on the corpus development, word segmentation, and word embeddings, respectively.
\par \setlength{\parindent}{3ex}
The power of word embeddings in NLP  was empirically estimated by proposing a neural language model~\cite{bengio2003neural} and multitask learning~\cite{collobert2008unified}, but recently usage of word embeddings in deep neural algorithms has become integral element~\cite{bojanowski2017enriching} for performance acceleration in deep NLP applications.
 The CBoW and SG~\cite{mikolov2013efficient}~\cite{mikolov2013distributed} popular word2vec neural architectures yielded high quality vector representations in lower computational cost with integration of character-level learning on large corpora in terms of semantic and syntactic word similarity later extended~\cite{bojanowski2017enriching}~\cite{mikolov2018advances}. Both approaches produce state-of-the-art accuracy with fast training performance, better representations of less frequent words and efficient representation of phrases as well.
~\cite{qiu2014co} proposed NN based approach for generating morphemic-level word embeddings, which surpassed all the existing embedding models in intrinsic evaluation. A count-based GloVe model~\cite{pennington2014glove} also yielded state-of-the-art results in an intrinsic evaluation and downstream NLP tasks. 
\par \setlength{\parindent}{3ex} 
The performance of Word embeddings is evaluated using intrinsic~\cite{schnabel2015evaluation}~\cite{pierrejean2018towards}  and extrinsic evaluation~\cite{nayak2016evaluating} methods. The performance of word embeddings can be measured with intrinsic and extrinsic evaluation approaches. The intrinsic approach is used to measure the internal quality of word embeddings such as querying nearest neighboring words and calculating the semantic or syntactic similarity between similar word pairs. A method of direct comparison for intrinsic evaluation of word embeddings measures the neighborhood of a query word in vector space. The key advantage of that method is to reduce bias and create insight to find data-driven relevance judgment. An extrinsic evaluation approach is used to evaluate the performance in downstream NLP tasks, such as parts-of-speech tagging or named-entity recognition~\cite{schnabel2015evaluation}, but the Sindhi language lacks annotated corpus for such type of evaluation. Moreover, extrinsic evaluation is time consuming and difficult to interpret. Therefore, we \textbf{opt} intrinsic evaluation method~\cite{nayak2016evaluating} to get a quick insight into the quality of proposed Sindhi word embeddings by measuring the cosine distance between similar words and using WordSim353 dataset. A study reveals that the choice of optimized hyper-parameters~\cite{levy2015improving} has a great impact on the quality of pretrained word embeddings as compare to desing a novel algorithm. Therefore, we optimized the hyperparameters for generating robust Sindhi word embeddings using CBoW, SG and GloVe models. The embedding visualization is also useful to visualize the similarity of word clusters. Therefore, we use t-SNE~\cite{maaten2008visualizing} dimensionality reduction algorithm for compressing high dimensional embedding into 2-dimensional $x$,$y$ coordinate pairs with PCA~\cite{lebret2013word}. The PCA is useful to combine input features by dropping the least important features while retaining the most valuable features.
\begin{table}[hptp]
    \centering
    \begin{tabular}{lll}
    \hline
         Paper & Related works &   Resource  \\
    \hline
   ~\cite{grave2018learning} & Word embedding & Wiki-dumps (2016)\\
  ~\cite{rahman2010towards} & Text Corpus &  4.1M tokens   \\
  ~\cite{motlani2016developing} & Corpus development & Wiki-dumps (2016) \\
  ~\cite{dootio2018unicode} & Labelled corpus & 6.8K   \\
  ~\cite{dootio2019development} & Sentiment analysis & 31.5K tokens \\
  ~\cite{bhatti2014word} & Text Segmentation &  1575K  \\
  Proposed work & Raw Corpus  & 61.39 M tokens \\
            & Word embeddings  & 61.39M tokens \\
    \hline
    \end{tabular}  
    \caption{Comparison of existing and proposed work on Sindhi corpus construction and word embeddings}
    \label{tab:related_work}
\end{table}
\section{Methodology}
\label{sec:methodology}
This section presents the employed methodology in detail for corpus acquisition, preprocessing, statistical analysis, and generating Sindhi word embeddings.
\subsection{Task description}
\label{sec:task_description}
We initiate this work from scratch by collecting large corpus from multiple web resources. After preprocessing and statistical analysis of the corpus, we generate Sindhi word embeddings with state-of-the-art CBoW, SG, and GloVe algorithms. The generated word embeddings are evaluated using the intrinsic evaluation approaches of cosine similarity between nearest neighbors, word pairs, and WordSim-353 for distributional semantic similarity. Moreover, we use t-SNE with PCA for the comparison of the distance between similar words via visualization.  
\subsection{Corpus acquisition}
\label{sec:corpus_acquisition}
The corpus is a collection of human language text~\cite{schafer2013web} built with a specific purpose. However, the statistical analysis of the corpus provides quantitative, reusable data, and an opportunity to examine intuitions and ideas about language. Therefore, the corpus has great importance for the study of written language to examine the text. In fact, realizing the necessity of large text corpus for Sindhi, we started this research by collecting raw corpus from multiple web resource using web-scrappy framwork\footnote{\url{https://github.com/scrapy/scrapy}} for extraction of news columns of daily Kawish\footnote{\url{http://kawish.asia/Articles1/index.htm}} and Awami Awaz\footnote{\url{http://www.awamiawaz.com/articles/294/}} Sindhi newspapers, Wikipedia dumps\footnote{\url{https://dumps.wikimedia.org/sdwiki/20180620/}}, short stories and sports news from Wichaar\footnote{\url{http://wichaar.com/news/134/}, accessed in Dec-2018} social blog, news from Focus Word press blog\footnote{\url{https://thefocus.wordpress.com/} accessed in Dec-2018}, historical writings, novels, stories, books from Sindh Salamat\footnote{\url{http://sindhsalamat.com/}, accessed in Jan-2019} literary websites, novels, history and religious books from Sindhi Adabi Board \footnote{\url{http://www.sindhiadabiboard.org/catalogue/History/Main_History.HTML}} 
and tweets regarding news and sports are collected from twitter\footnote{\url{https://twitter.com/dailysindhtimes}}.
\subsection{Preprocessing} \label{sec:preprocessing}
\begin{figure}
	\centering
	\includegraphics[width=13cm, height=3.5cm]{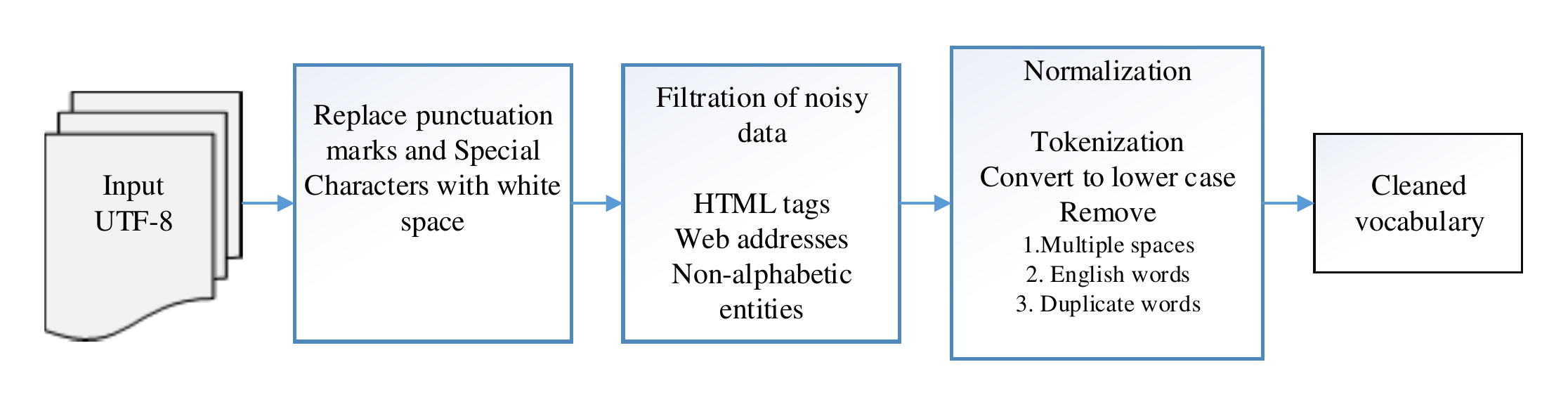}
	\caption{Employed preprocessing pipeline for text cleaning}
	\label{fig:preprocessing}
\end{figure}

The preprocessing of text corpus obtained from multiple web resources is a challenging task specially it becomes more complicated when working on low-resourced language like Sindhi due to the lack of open-source preprocessing tools such as NLTK~\cite{loper2002nltk} for English. Therefore, we design a preprocessing pipeline depicted in Figure \ref{fig:preprocessing} for the filtration of unwanted data and vocabulary of other languages such as English to prepare input for word embeddings. Whereas, the involved preprocessing steps are described in detail below the Figure \ref{fig:preprocessing}. Moreover, we reveal the list of Sindhi stop words~\cite{pandey2009evaluating} which is labor intensive and requires human judgment as well. Hence, the most frequent and least important words are classified as stop words with the help of a Sindhi linguistic expert. The partial list of Sindhi stop words is given in \ref{tab:stopwords}. We use python programming language for designing the preprocessing pipeline using regex and string functions.
\begin{itemize}
	\item \textbf{Input:} The collected text documents were concatenated for the input in UTF-8 format.
	\item \textbf{Replacement symbols:} The punctuation marks of a full stop, hyphen, apostrophe, comma, quotation, and exclamation marks replaced with white space for authentic tokenization because without replacing these symbols with white space the words were found joined with their next or previous corresponding words.  
	\item \textbf{Filtration of noisy data:} The text acquisition from web resources contain a huge amount of noisy data. Therefore, we filtered out unimportant data such as the rest of the punctuation marks, special characters, HTML tags, all types of numeric entities, email, and web addresses.  
	\item \textbf{Normalization:} In this step, We tokenize the corpus then normalize to lower-case for the filtration of multiple white spaces, English vocabulary, and duplicate words. The stop words were only filtered out for preparing input for GloVe. However, the sub-sampling approach in  CBoW and SG can discard most frequent or stop words automatically.
\end{itemize}
\subsection{Word embedding models} \label{sec:embedding_models}
The NN based approaches have produced state-of-the-art performance in NLP with the usage of robust word embedings generated from the large unlabelled corpus. Therefore, word embeddings have become the main component for setting up new benchmarks in NLP using deep learning approaches. Most recently, the use cases of word embeddings are not only limited to boost statistical NLP applications but can also be used to develop language resources such as automatic construction of WordNet~\cite{khodak2017automated} using the unsupervised approach.  
\par \setlength{\parindent}{3ex}
The word embedding can be precisely defined as the encoding of vocabulary $V$ into $N$ and the word $w$ from $V$ to vector $\overrightarrow{w} $ into $N$-dimensional embedding space. They can be broadly categorized into predictive and count based methods, being generated by employing co-occurrence statistics, NN algorithms, and probabilistic models. The GloVe~\cite{pennington2014glove} algorithm treats each word as a single entity in the corpus and generates a vector of each word. However, CBoW and SG~\cite{mikolov2013efficient}~\cite{mikolov2013distributed}, later extended~\cite{bojanowski2017enriching}~\cite{mikolov2018advances}, well-known as word2vec rely on simple two layered NN architecture which uses linear activation function in hidden layer and softmax in the output layer. The work2vec model treats each word as a bag-of-character n-gram.
\subsection{GloVe} \label{sec:glove}
The GloVe is a log-bilinear regression model~\cite{pennington2014glove} which combines two methods of local context window and global matrix factorization for training word embeddings of a given vocabulary in an unsupervised way. It weights the contexts using the harmonic function, for example, a context word four tokens away from an occurrence will be counted as $\frac{1}{4}$. The Glove’s implementation represents word $\boldsymbol{w} \in \boldsymbol{V}_{\boldsymbol{w}}$ and context $\boldsymbol{c} \in \boldsymbol{V}_{c}$ in $D$-dimensional vectors $\overrightarrow{w}$ and $\overrightarrow{c}$ in a following way, 
\begin{equation}
\overrightarrow{w} \cdot \overrightarrow{c}+b_{w}+b_{c}=\log (\#(w, c)) \forall(w, c) \in D
\end{equation}
Where $\boldsymbol{b}_{\boldsymbol{w}}$ and $b_{c}$ represent word and context biases also to be learned as parameters of $\overrightarrow{w} $ and $\overrightarrow{c} $. The objective of GloVe algorithm is to learn word embeddings by taking the log-count matrix~\cite{levy2015improving} shifted by the bias terms of entire vocabulary as,
\begin{equation}
 M^{\log (\#(\boldsymbol{w}, c))} \approx \boldsymbol{W} \cdot \boldsymbol{C}^{T}+\boldsymbol{b}^{\overrightarrow{w}}+\boldsymbol{b}^{\overrightarrow{c}} 
\end{equation}
where, $\boldsymbol{b}^{\overrightarrow{w}}$  is row vector $\left|V_{w}\right|$ and $\boldsymbol{b}^{\overrightarrow{c}}$ is $\left|V_{c}\right|$  is column vector. 
\subsection{Continuous bag-of-words} \label{CBoW}
The standard CBoW is the inverse of SG~\cite{mikolov2013efficient} model, which predicts input word on behalf of the context. The length of input in the CBoW model depends on the setting of context window size which determines the distance to the left and right of the target word. Hence the context is a window that contain neighboring words such as by giving  $w=\left\{w_{1}, w_{2}, \dots \dots w_{t}\right\}$ a sequence of words $T$, the objective of the CBoW is to maximize the probability of given neighboring words such as, 
\begin{equation}
    \sum_{t=0}^{T} \log p\left(w_{t} | c_{t}\right)
\end{equation}
Where $c_{t}$ is context of  $t^{\text {th}}$ word for example with window $w_{t-c}, \ldots . . w_{t-1}, w_{t+1} \ldots w_{t+c}$ of size $2 c$.
\subsection{Skip gram} \label{sec:skipgram}
The SG model predicts surrounding words by giving input word~\cite{mikolov2013distributed} with training objective of learning good word embeddings that efficiently predict the neighboring words. The goal of skip-gram is to maximize average log-probability of words $w=\left\{w_{1}, w_{2}, \dots \dots w_{t}\right\}$  across the entire training corpus,
\begin{equation}
\mathrm{J}(\theta) \frac{1}{\mathrm{T}} \sum_{t=0}^{T}\left(\sum_{ -c \leq j \leq c, j} \log p\left(\mathrm{w}_{\mathrm{t}+j} | c_{t}\right)\right)
\end{equation}
Where $c_{t}$ denotes the context of words indices set of nearby $w_{t}$ words in the training corpus.
\subsection{Hyperparameters} \label{subsampling}
\subsubsection{Sub-sampling}
\label{sec:sub_sampling}
Th sub-sampling~\cite{mikolov2013distributed} approach is useful to dilute most frequent or stop words, also accelerates learning rate, and increases accuracy for learning rare word vectors. Numerous words in English, e.g., ‘the’, ‘you’, ’that’ do not have more importance, but these words appear very frequently in the text. However, considering all the words equally would also lead to over-fitting problem of model parameters~\cite{mikolov2018advances} on the frequent word embeddings and under-fitting on the rest. Therefore, it is useful to count the imbalance between rare and repeated words. The sub-sampling technique randomly removes most frequent words with some threshold $t$ and probability $p$ of words and frequency $f$ of words in the corpus.
\begin{equation}
    P\left(w_{i}\right)=1-\sqrt{\frac{t}{f\left(w_{i}\right)}}
\end{equation}
Where each word$w_{i}$ is discarded with computed probability in training phase,  $f(w_i )$ is frequency of word $w_{i}$ and $t>0$ are parameters.
\subsubsection{Dynamic context window}
\label{sec:context_window}
The traditional word embedding models usually use a fixed size of a context window. For instance, if the window size \textit{ws=6}, then the target word apart from 6 tokens will be treated similarity as the next word. The scheme is used to assign more weight to closer words, as closer words are generally considered to be more important to the meaning of the target word. The CBoW, SG and GloVe models employ this weighting scheme. The GloVe model weights the contexts using a harmonic function, for example, a context word four tokens away from an occurrence will be counted as $\frac{1}{4}$. However, CBoW and SG implementation equally consider the contexts by dividing the \textit{ws} with the distance from target word, e.g. \textit{ws=6} will weigh its context by $\frac{6}{6} \frac{5}{6} \frac{4}{6} \frac{3}{6} \frac{2}{6} \frac{1}{6}$
\subsubsection{Sub-word model}
\label{sec:sub_word}
The sub-word model~\cite{mikolov2018advances} can learn the internal structure of words by sharing the character representations across words. In that way, the vector for each word is made of the sum of those character $n-gram$. Such as, a vector of a word “table” is a sum of $n-gram$ vectors by setting the letter $n-gram$ size $min=3$ to $max=6$  as, $<ta, tab, tabl, table, table>, abl, able, able>, ble, ble>, le>$, we can get all sub-words of "table" with minimum length of $minn=3$ and maximum length of $maxn=6$.  The $<$ and $>$ symbols are used to separate prefix and suffix words from other character sequences. In this way, the sub-word model utilizes the principles of morphology, which improves the quality of infrequent word representations. In addition to character $n-grams$, the input word $w$ is also included in the set of character  $n-gram$, to learn the representation of each word. We obtain scoring function using a input dictionary of $n-grams$ with size $K$ by giving word $w$ , where $K_{w} \subset\{1, \ldots, K\}$. A word representation $Z_{k}$ is associated to each $n-gram$ $Z$.  Hence, each word is represented by the sum of character $n-gram$ representations, where, $s$  is the scoring function in the following equation,
\begin{equation}
    s(w, c)=\sum_{k \in K_{j}} z_{k}^{T} v c
\end{equation}
\subsubsection{Position-dependent weights}
\label{sec:position_dependent_weights}
The position-dependent weighting approach~\cite{mnih2013learning} is used to avoid direct encoding of representations for words and their positions which can lead to over-fitting problem. The approach learns positional representations in contextual word representations and used to reweight word embedding. Thus, it captures good contextual representations at lower computational cost,
\begin{equation}
    \boldsymbol{v}_{C} \boldsymbol{V}=\sum_{p \in P} \boldsymbol{d}_{p} \odot \boldsymbol{u}_{t+p}
\end{equation}
Where $p$ is individual position in context window associated with $d_{p}$ vector.  Afterwards the context vector reweighted by their positional vectors is average of context words. The relative positional set is  $P$ in context window and $v_{C}$ is context vector of $w_{t}$ respectively.  
\subsubsection{Shifted point-wise mutual information}
\label{SPWMI}
The use sparse Shifted Positive Point-wise Mutual Information (SPPMI)~\cite{levy2014neural} word-context matrix in learning word representations improves results on two word similarity tasks.
The CBoW and SG have $k$ (number of negatives)~\cite{mikolov2013efficient}~\cite{mikolov2013distributed} hyperparameter, which affects the value that both models try to optimize for each $(w, c): P M I(w, c)-\log k$. Parameter  $k$  has two functions of better estimation of negative examples, and it performs as before observing the probability of positive examples (actual occurrence of $w,c$). 
\subsubsection{Deleting rare words}
\label{sec:delet_rare_words}
Before creating a context window, the automatic deletion of rare words also leads to performance gain in CBoW, SG and GloVe models, which further increases the actual size of context windows.
\subsection{Evaluation methods}
\label{sec:evaluation_matrix}
The intrinsic evaluation is based on semantic similarity~\cite{schnabel2015evaluation}  in word embeddings. The word similarity measure approach states~\cite{levy2015improving} that the words are similar if they appear in the similar context. We measure word similarity of proposed Sindhi word embeddings using dot product method and WordSim353.
\subsubsection{Cosine similarity}
\label{cosine_similarity}
The cosine similarity between two non-zero vectors is a popular measure that calculates the cosine of the angle between them which can be derived by using the Euclidean dot product method. The dot product is a multiplication of each component from both vectors added together. The result of a dot product between two vectors isn’t another vector but a single value or a scalar. The dot product for two vectors can be defined as:  $\overrightarrow{a}=\left(a_{1}, a_{2}, a_{3}, \dots, a_{n}\right)$ and $\overrightarrow{b}=\left({b}_{1}, {b}_{2}, {b}_{3}, \ldots, {b}_{n}\right)$ where $a_{n}$ and $b_{n}$ are the components of the vector and $n$ is dimension of vectors such as,
\begin{equation}
    \overrightarrow{{a}} \cdot \overrightarrow{b}=\sum_{i=1}^{n} a_{i} {b}_{i=} a_{1} {b}_{1}+{a}_{2} {b}_{2}+\ldots \ldots \ldots \ldots a_{n} {b}_{n}
\end{equation}
However, the cosine of two non-zero vectors can be derived by using the Euclidean dot product formula,
\begin{equation}
    \overrightarrow{a} \cdot \overrightarrow{b}=\|\overrightarrow{a}\|\|\overrightarrow{b}\| \cos ({\theta})
\end{equation}
Given $a_{i}$ two vectors of attributes $a$ and $b$, the cosine similarity, $\cos ({\theta})$, is represented using a dot product and magnitude as,
\begin{equation}
    \text { similarity }=\cos (\theta)=\frac{\bar{a} \cdot \vec{b}}{\|\vec{a}\|\|\vec{b}\|} \frac{\sum_{i=1}^{n} a_{i} b_{i}}{\sqrt{\sum_{i=1}^{n} a_{1}^{2}} \sqrt{\sum_{i=1}^{n} b_{1}^{2}}}
    \label{cosine similarity}
\end{equation}
where $a_{i}$ and $b_{i}$ are components of vector $\overrightarrow{a}$  and $\overrightarrow{b}$, respectively.
\subsubsection{WordSim353}
\label{sec:wordsim353}
The WordSim353~\cite{finkelstein2002placing} is popular for the evaluation of lexical similarity and relatedness. The similarity score is assigned with 13 to 16 human subjects with semantic relations~\cite{agirre2009study}  for 353 English noun pairs.  Due to the lack of annotated datasets in the Sindhi language, we translated WordSim353 using English to Sindhi bilingual dictionary\footnote{\url{http://dic.sindhila.edu.pk/index.php?txtsrch=}} for the evaluation of our proposed Sindhi word embeddings and SdfastText.  We use the Spearman correlation coefficient for the semantic and syntactic similarity comparison which is used to used to discover the strength of linear or nonlinear relationships if there are no repeated data values.  A perfect Spearman’s correlation of $+1$ or $-1$ discovers the strength of a link between two sets of data (word-pairs)  when observations are monotonically increasing or decreasing functions of each other in a following way:
\begin{equation}
    r_{s}=1-\frac{6 \sum_{i}^{n} d_{i}^{2}}{n\left(n^{2}-1\right)}
    \label{Spearman_corr}
\end{equation}
where $r_s$ is the rank correlation coefficient, $n$ denote the number of observations, and $d^i$ is the rank difference between $i^{th}$ observations.
\section{Statistical analysis of corpus}
\label{sec:statistical_analysis_of_corpus}
The large corpus acquired from multiple resources is rich in vocabulary. We present the complete statistics of collected corpus (see Table \ref{tab:dataset_statistics} with number of sentences, words and unique tokens. 
\begin{table}
	\centering
	\begin{tabular}{lllll}
		\hline
		\textbf{Source} &  \textbf{Category}  &  \textbf{Sentences} &      \textbf{Vocabulary} & \textbf{Unique words} \\
		\hline
		Kawish & News columns  &    473,225 & 13,733,379 &    109,366 \\
		\hline
		Awami awaz & News columns &    107,326 &  7,487,319 &     65,632 \\
		\hline
		Wikipedia & Miscellaneous &    844,221 &  8,229,541 &    245,621 \\
		Social Blogs & Stories, sports  &      7,018 &    254,327 &     10,615 \\
		& History, News &      3,260 &    110,718 &      7,779 \\
		\hline
		Focus word press & Short Stories &     63,251 &    968,639 &     28,341 \\
		&     Novels &     36,859 &    998,690 &     18,607 \\
		&  Safarnama &    138,119 &  2,837,595 &     53,193 \\
		\hline
		Sindh Salamat &    History &    145,845 &  3,493,020 &     61,993 \\
		&   Religion &     96,837 &  2,187,563 &     39,525 \\
		&    Columns &     85,995 &  1,877,813 &     33,127 \\
		& Miscellaneous  &    719,956 &  9,304,006 &    168,009 \\
		\hline
		Sindhi Adabi Board & History books &    478,424 &  9,757,844 &     57,854 \\
		Twitter  & News tweets &     10,752 &    159,130 &      9,794 \\
		\hline
		Total &            &  3,211,088 & 61,399,584 &     908,456 \\
		\hline
	\end{tabular}
	\caption{Complete statistics of collected corpus from multiple resources}
	\label{tab:dataset_statistics}
\end{table}
\subsection{Letter occurrences} \label{sec:letter_occurances}
\begin{figure}
	\centering
	\includegraphics[width=15cm]{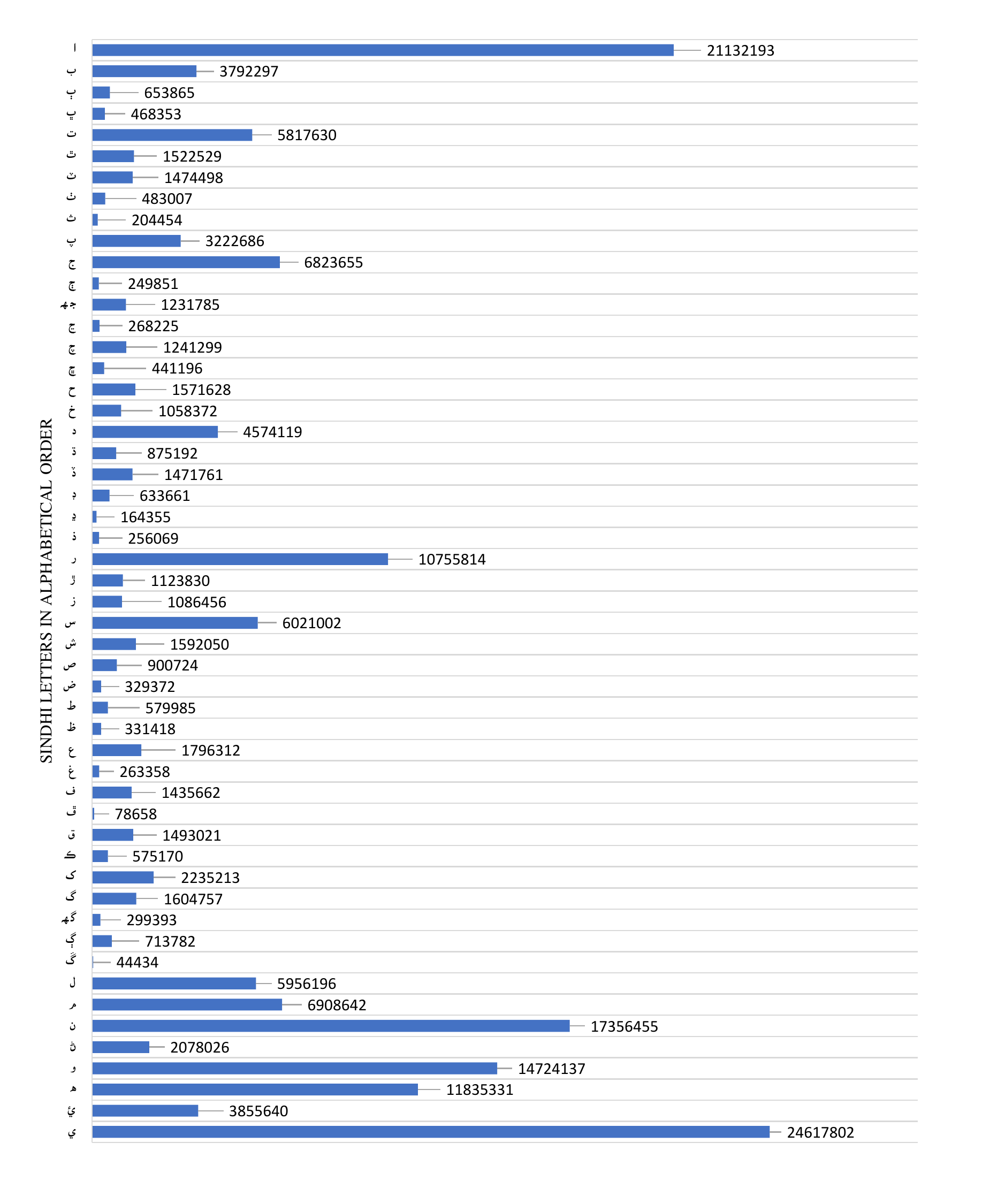}
	\caption{Frequency distribution of letter occurrences}
	\label{fig:letter_freq}
\end{figure}

The frequency of letter occurrences in human language is not arbitrarily organized but follow some specific rules which enable us to describe some linguistic regularities. The Zipf’s law~\cite{corral2015zipf} suggests that if the frequency of letter or word occurrence ranked in descending order such as,  
\begin{equation}
F_{r} = \frac{a}{r^{b}}
\end{equation}
Where, $F_{r}$ is the letter frequency of r$^{th}$ rank, $a$ and $b$ are parameters of input text.
The comparative letter frequency in the corpus is the total number of occurrences of a letter divided by the total number of letters present in the corpus. The letter frequencies in our developed corpus are depicted in Figure \ref{fig:letter_freq}; however, the corpus contains 187,620,276 total number of the character set. Sindhi Persian-Arabic alphabet consists of 52 letters but in the vocabulary 59 letters are detected, additional seven letters are modified uni-grams and standalone honorific symbols.
\subsection{Letter n-grams frequency} \label{sec:2}
We denote the combination of letter occurrences in a word as n-grams, where each letter is a gram in a word. The letter n-gram frequency is carefully analyzed in order to find the length of words which is essential to develop NLP systems, including learning of word embeddings such as choosing the minimum or maximum length of sub-word for character-level representation learning~\cite{mikolov2018advances}. We calculate the letter n-grams in words along with their percentage in the developed corpus (see Table \ref{tab:letter_ngrams}). The bi-gram words are most frequent, mostly consists of stop words and secondly, 4-gram words have a higher frequency. 
\begin{table}
	\centering
	\begin{tabular}{lll}
		\hline
		n-grams  & Frequency  & \% in corpus \\
		\hline
		Uni-gram &       936,301 &    1.52889 \\
		Bi-gram &    19,187,314 &    31.3311 \\
		Tri-gram &   11,924,760 &     19.472 \\
		4-gram &     14,334,444 &    23.4068 \\
		5-gram &     9,459,657 &    15.4467 \\
		6-gram &     3,347,907 &     5.4668 \\
		7-gram &     1,481,810 &     2.4196 \\
		8-gram &       373,417 &     0.6097 \\
		9-gram &       163,301 &     0.2666 \\
		10-gram &      21,287 &     0.0347 \\
		11-gram &      5,892 &     0.0096 \\
		12-gram &       3,033 &     0.0049 \\
		13-gram &       1,036 &     0.0016 \\
		14-gram &        295 &     0.0004 \\
		\hline
		Total &     61,240,454 &        100 \\
		\hline
	\end{tabular}  
	\caption{Length of letter n-grams in words, distinct words, frequency and percentage in corpus}
	\label{tab:letter_ngrams}
\end{table}
\subsection{Word Frequencies}
The word frequency count is an observation of word occurrences in the text. The commonly used words are considered to be with higher frequency, such as the word “the" in English. Similarly, the frequency of rarely used words to be lower. Such frequencies can be calculated at character or word-level. We calculate word frequencies by counting a word $w$ occurrence in the corpus $c$, such as,  
\begin{equation}
    \text { freq}(w)=\sum_{\mathrm{k}=0}^{\mathrm{k}} w_{k} \in c
    \label{word_freq}
\end{equation}
Where the frequency of $w$ is the sum of every occurrence $k$ of $w$ in $c$. 
\subsection{Stop words}
\label{sec:stop_words}
\begin{figure}
	\centering
	\includegraphics[width=14cm]{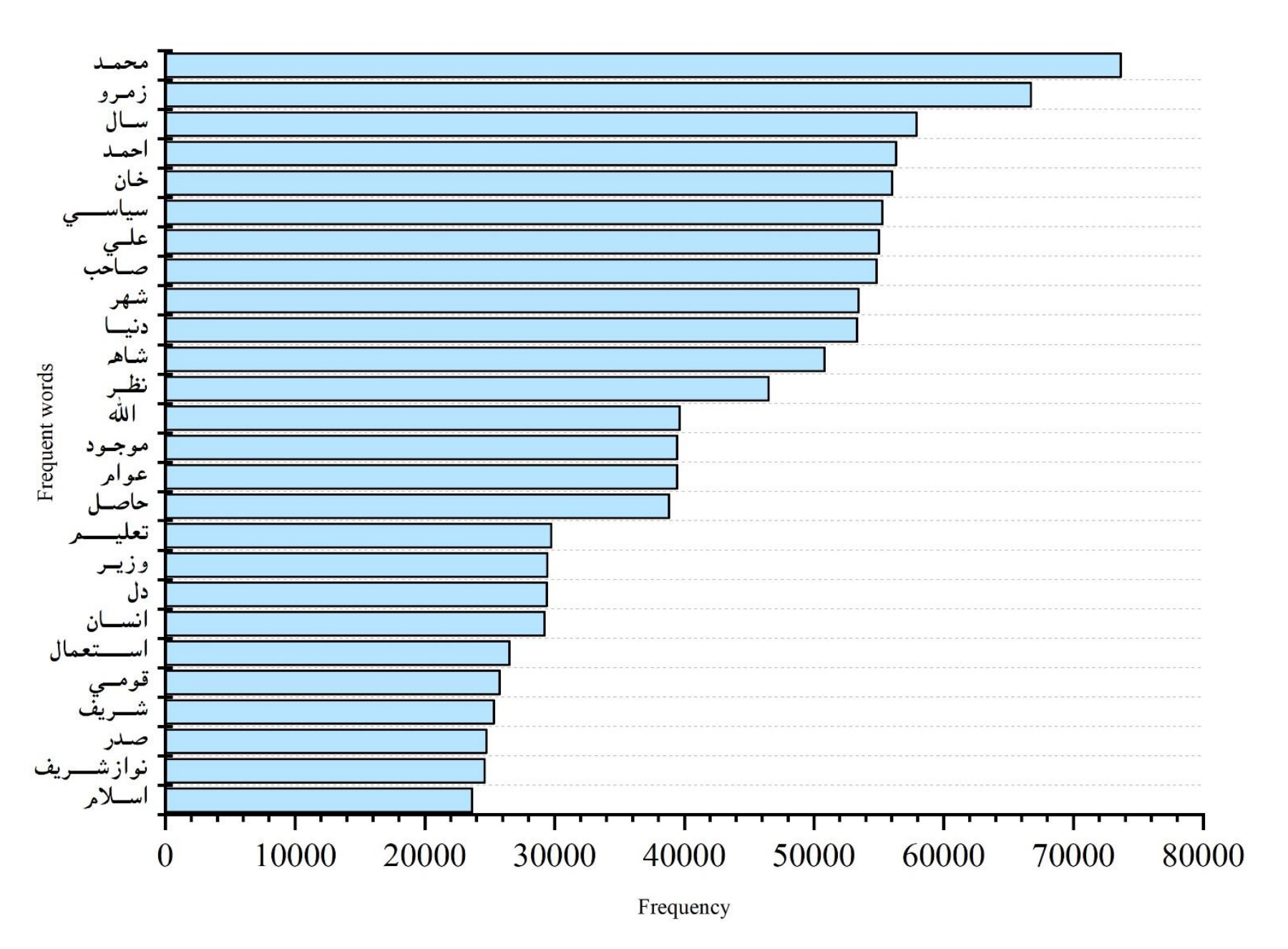}
	\caption{Most frequent words after filtration of stop words.}
	\label{fig:freq_words}
\end{figure}

The most frequent and least important words in NLP are often classified as stop words. The removal of such words can boost the performance of the NLP model~\cite{pandey2009evaluating}, such as sentiment analysis and text classification. But the construction of such words list is time consuming and requires user decisions. Firstly, we determined Sindhi stop words by counting their term frequencies using Eq.~\ref{word_freq}, and secondly, by analysing their grammatical status with the help of Sindhi linguistic expert because all the frequent words are not stop words (see Figure \ref{fig:freq_words}). After determining the importance of such words with the help of human judgment, we placed them in the list of stop words. The total number of detected stop words is 340 in our developed corpus. The partial list of most frequent Sindhi stop words is depicted in Table \ref{tab:stopwords} along with their frequency. The filtration of stop words is an essential preprocessing step for learning GloVe~\cite{pennington2014glove} word embeddings; therefore, we filtered out stop words for preparing input for the GloVe model. However, the sub-sampling approach ~\cite{bojanowski2017enriching}~\cite{mikolov2018advances} is used to discard such most frequent words in CBoW and SG models.
\begin{table}
	\centering
	\includegraphics{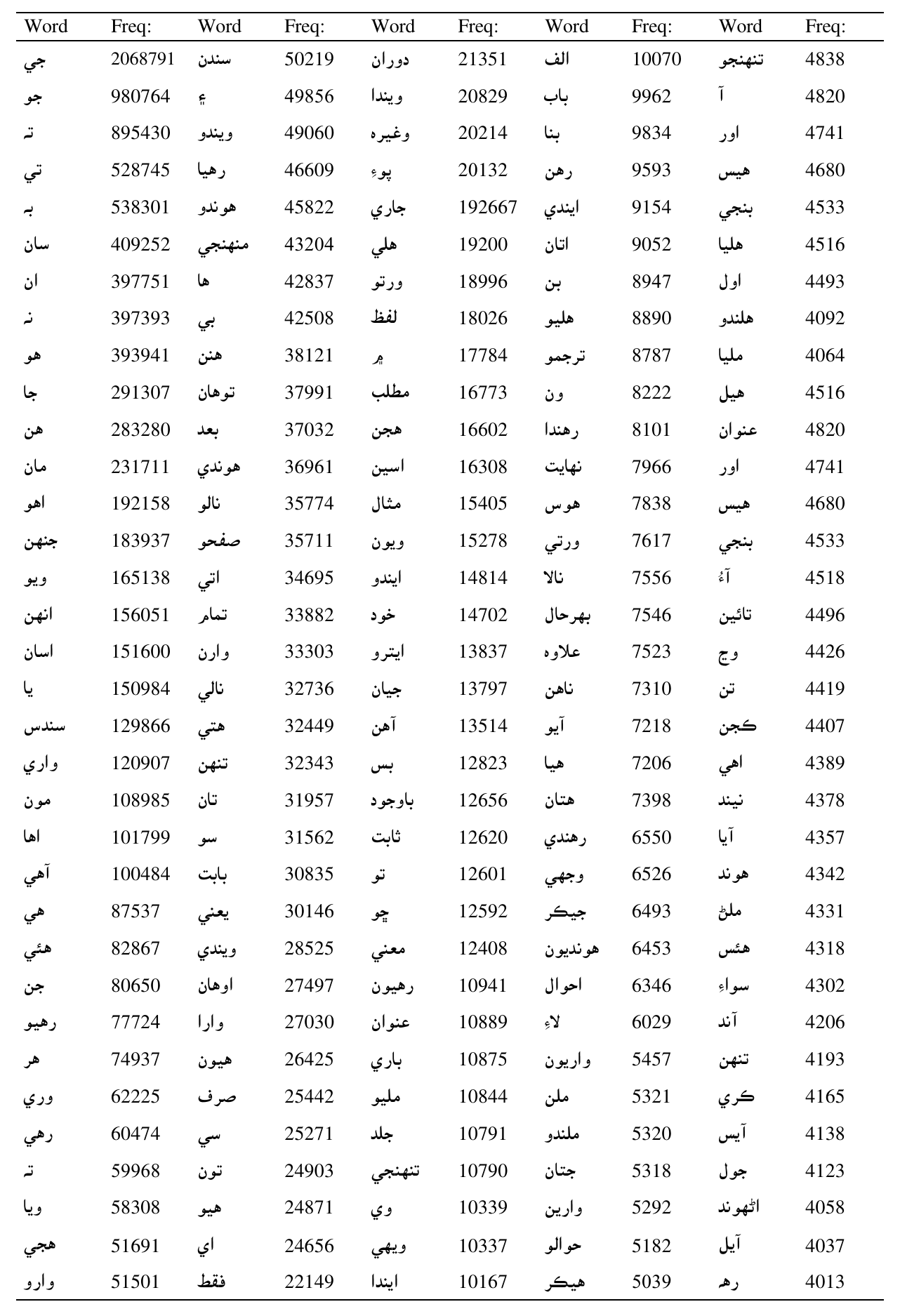}
	\caption{Partial list of  most frequent Sindhi stop words along with frequency in the developed corpus.}
	\label{tab:stopwords}
\end{table}
\section{Experiments and results}
\label{sec:exp_set}
Hyperparameter optimization~\cite{schnabel2015evaluation}is more important than designing a novel algorithm. We carefully choose to optimize the dictionary and algorithm-based parameters of CBoW, SG and GloVe algorithms. Hence, we conducted a large number of experiments for training and evaluation until the optimization of most suitable hyperparameters depicted in Table \ref{tab:emb_parameters} and discussed in Section~\ref{Hyperpara_optimization}. The choice of optimized hyperparameters is based on The high cosine similarity score in retrieving nearest neighboring words, the semantic, syntactic similarity between word pairs, WordSim353, and visualization of the distance between twenty nearest neighbours using t-SNE respectively. All the experiments are conducted on GTX 1080-TITAN GPU. 
\subsection{Hyperparameter optimization}
\label{Hyperpara_optimization}
The state-of-the-art SG, CBoW~\cite{mikolov2013efficient}~\cite{bojanowski2017enriching}~\cite{mikolov2013distributed}~\cite{mikolov2018advances} and Glove~\cite{pennington2014glove} word embedding algorithms are evaluated by parameter tuning for development of Sindhi word embeddings. These parameters can be categories into dictionary and algorithm based, respectively. The integration of character n-gram in learning word representations is an ideal method especially for rich morphological languages because this approach has the ability to compute rare and misspelled words. Sindhi is also a rich morphological language. Therefore more robust embeddings became possible to train with the hyperparameter optimization of SG, CBoW and GloVe algorithms.  We tuned and evaluated the hyperparameters of three algorithms individually which are discussed as follows:
\begin{table}
	\centering
	\begin{tabular}{lll}
		\hline
		Parameter  & CBoW, SG   &  GloVe \\
		\hline
		Epoch   &   40     &  40 \\
		lr      &  0.25    &    0.25 \\
		$D$     &   300    &     300 \\
		minn char & 02    &    -- \\
		maxn char & 07   &     -- \\
		ws      &   7      &    7 \\
		NS      &   20  &     -- \\
		minw    &   4   &     4 \\
		\hline
	\end{tabular}  
	\caption{Optimized parameters for CBoW, SG and GloVe models }
	\label{tab:emb_parameters}
\end{table}
\begin{itemize}
	\item \textbf{Number of Epochs:} Generally, more epochs on the corpus often produce better results but more epochs take long training time. Therefore, we evaluate $10$, $20$, $30$ and $40$ epochs for each word embedding model, and $40$ epochs constantly produce good results.
	\item \textbf{Learning rate (lr):} We tried lr of $0.05$, $0.1$, and $0.25$, the optimal lr $(0.25)$ gives the better results for training all the embedding models.
	\item \textbf{Dimensions ($D$):} We evaluate and compare the quality of $100-D$, $200-D$, and  $300-D$  using WordSim353 on different $ws$, and the optimal  $300-D$  are evaluated with cosine similarity matrix for querying nearest neighboring words and calculating the similarity between word pairs. The embedding dimensions have little affect on the quality of the intrinsic evaluation process. However, the selection of embedding dimensions might have more impact on the accuracy in certain downstream NLP applications. The lower embedding dimensions are faster to train and evaluate.
	\item \textbf{Character n-grams:} The selection of minimum (minn) and the maximum (maxn) length of character $n-grams$ is an important parameter for learning character-level representations of words in CBoW and SG models. Therefore, the n-grams from $3-9$ were tested to analyse the impact on the accuracy of embedding. We optimized the length of character n-grams from $minn=2$ and $maxn=7$ by keeping in view the word frequencies depicted in Table \ref{tab:letter_ngrams}.
	\item \textbf{Window size (ws):} The large \textit{ws} means considering more context words and similarly less \textit{ws} means to limit the size of context words. By changing the size of the dynamic context window, we tried the \textit{ws} of 3, 5, 7 the optimal \textit{ws=7} yield consistently better performance.
	\item \textbf{Negative Sampling (NS):} : The more negative examples yield better results, but more negatives take long training time. We tried 10, 20, and 30 negative examples for CBoW and SG. The best negative examples of 20 for CBoW and SG significantly yield better performance in average training time.
	\item \textbf{Minimum word count (minw):} We evaluated the range of minimum word counts from 1 to 8 and analyzed that the size of input vocabulary is decreasing at a large scale by ignoring more words similarly the vocabulary size was increasing by considering rare words. Therefore, by ignoring words with a frequency of less than 4 in CBoW, SG, and GloVe consistently yields better results with the vocabulary of 200,000 words.
	\item \textbf{Loss function (ls):}   we use hierarchical softmax (hs) for CBoW, negative sampling (ns) for SG and default loss function for GloVe~\cite{pennington2014glove}.
	\item The recommended verbosity level, number of buckets, sampling threshold, number of threads are used for training CBoW, SG~\cite{mikolov2018advances}, and GloVe~\cite{pennington2014glove}.
\end{itemize}
\section{Word similarity comparison of Word Embeddings}
\label{sec:word_similarity_comparision}
\begin{table}
	\centering
	\includegraphics{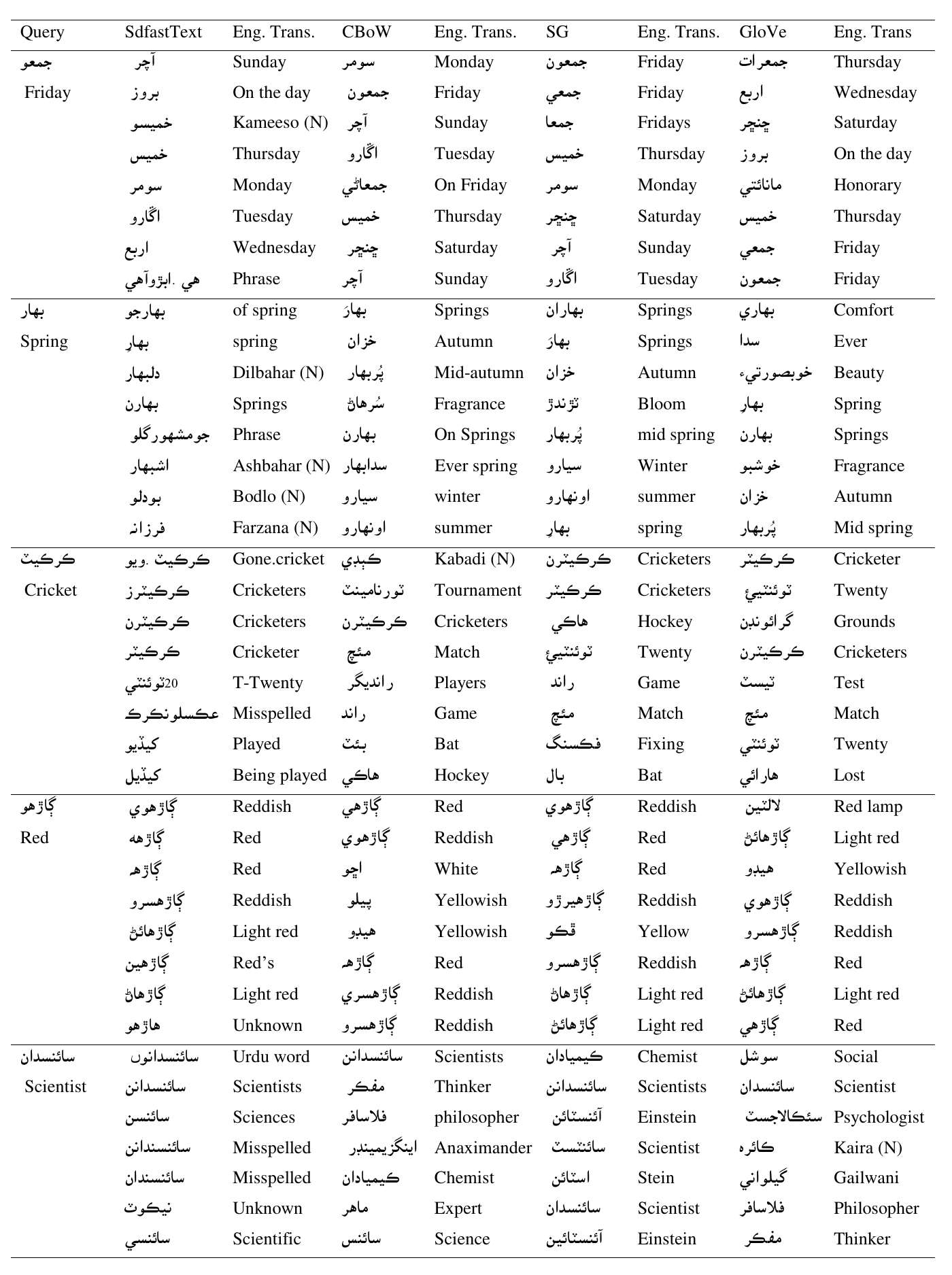}
	\caption{Eight nearest neighboring words of each query word with English translation}
	\label{tab:nearest_neighnors}
\end{table}
\subsection{Nearest neighboring words} \label{nearset_neighbors}
The cosine similarity matrix~\cite{levy2015improving} is a popular approach to compute the relationship between all embedding dimensions of their distinct relevance to query word. The words with similar context get high cosine similarity and geometrical relatedness to Euclidean distance, which is a common and primary method to measure the distance between a set of words and nearest neighbors. Each word contains the most similar top eight nearest neighboring words determined by the highest cosine similarity score using Eq.~\ref{cosine similarity}. We present the English translation of both query and retrieved words also discuss with their English meaning for ease of relevance judgment between the query and retrieved words.To take a closer look at the semantic and syntactic relationship captured in the proposed word embeddings, Table \ref{tab:nearest_neighnors} shows the top eight nearest neighboring words of five different query words {\textbf{Friday, Spring, Cricket, Red, Scientist}} taken from the vocabulary. As the first query word {\textit{Friday}} returns the names of days {\textit{Saturday, Sunday, Monday, Tuesday, Wednesday, Thursday}} in an unordered sequence. The SdfastText returns five names of days {\textit{Sunday, Thursday, Monday, Tuesday}} and {\textit{Wednesday}} respectively. The GloVe model also returns five names of days. However, CBoW and SG gave six names of days except {\textit{Wednesday}} along with different writing forms of query word {\textit{Friday}} being written in the Sindhi language which shows that CBoW and SG return more relevant words as compare to SdfastText and GloVe. The CBoW returned {\textit{Add}} and GloVe returns {\textit{Honorary}} words which are little similar to the querry word but SdfastText resulted two irrelevant words {\textit{Kameeso (N)}} which is a name (N) of person in Sindhi and {\textit{Phrase}} is a combination of three Sindhi words which are not tokenized properly. Similarly, nearest neighbors of second query word {\textit{Spring}} are retrieved accurately as names and seasons and semantically related to query word {\textit{Spring}} by CBoW, SG and Glove but SdfastText returned four irrelevant words of {\textit{Dilbahar (N), Pharase, Ashbahar (N)}} and {\textit{Farzana (N)}} out of eight. The third query word is {\textit{Cricket}}, the name of a popular game. The first retrieved word in CBoW is {\textit{Kabadi (N)}} that is a popular national game in Pakistan. Including {\textit{Kabadi (N)}} all the returned words by CBoW, SG and GloVe are related to {\textit{Cricket}} game or names of other games. But the first word in SdfastText contains a punctuation mark in retrieved word {\textit{Gone.Cricket}} that are two words joined with a punctuation mark (.), which shows the tokenization error in preprocessing step, sixth retrieved word {\textit{Misspelled}} is a combination of three words not related to query word, and {\textit{Played, Being played}} are also irrelevant and stop words. Moreover, fourth query word {\textit{Red}} gave results that contain names of closely related to query word and different forms of query word written in the Sindhi language. The last returned word {\textit{Unknown}} by SdfastText is irrelevant and not found in the Sindhi dictionary for translation. The last query word {\textit{Scientist}} also contains semantically related words by CBoW, SG, and GloVe, but the first {\textit{Urdu word}} given by SdfasText belongs to the Urdu language which means that the vocabulary may also contain words of other languages. Another {\textit{unknown}} word returned by SdfastText does not have any meaning in the Sindhi dictionary. More interesting observations in the presented results are the diacritized words retrieved from our proposed word embeddings and The authentic tokenization in the preprocessing step presented in Figure~\ref{fig:preprocessing}. However, SdfastText has returned tri-gram words of {\textit{Phrase}} in query words {\textbf{Friday, Spring}}, a {\textit{Misspelled}} word in {\textbf{Cricket}} and {\textbf{Scientist}} query words. Hence, the overall performance of our proposed SG, CBoW, and GloVe demonstrate high semantic relatedness in retrieving the top eight nearest neighbor words.
\subsection{Word pair relationship}
\label{sec:word_pair_relationship} 
Generally, closer words are considered more important to a word’s meaning. The word embeddings models have the ability to capture the lexical relations between words. Identifying such relationship that connects words is important in NLP applications. We measure that semantic relationship by calculating the dot product of two vectors using Eq.~\ref{cosine similarity}. The high cosine similarity score denotes the closer words in the embedding matrix, while less cosine similarity score means the higher distance between word pairs. We present the cosine similarity score of different semantically or syntactically related word pairs taken from the vocabulary in Table \ref{tab:word_pair_relationship} along with English translation, which shows the average similarity of 0.632, 0.650, 0.591 yields by CBoW, SG and GloVe respectively. The SG model achieved a high average similarity score of {\textbf{0.650}} followed by CBoW with a 0.632 average similarity score. The GloVe also achieved a considerable average score of 0.591 respectively. However, the average similarity score of SdfastText is 0.388 and the word pair {\textbf{Microsoft-Bill Gates}} is not available in the vocabulary of SdfastText. This shows that along with performance, the vocabulary in SdfastText is also limited as compared to our proposed word embeddings. %
\par\setlength{\parindent}{3ex} 
Moreover, the average semantic relatedness similarity score between countries and their capitals is shown in Table \ref{tab:country_capital} with English translation, where SG also yields the best average score of {\textbf{0.663}} followed by CBoW with 0.611 similarity score. The GloVe also yields better semantic relatedness of 0.576 and the SdfastText yield an average score of 0.391. The first query word {\textbf{China-Beijing}} is not available the vocabulary of SdfastText. However, the similarity score between {\textit{Afghanistan-Kabul}} is lower in our proposed CBoW, SG, GloVe models because the word {\textit{Kabul}} is the name of the capital of \textit{Afghanistan} as well as it frequently appears as an adjective in Sindhi text which means \textit{able}.
\begin{table}
	\centering
	\includegraphics[width=12 cm]{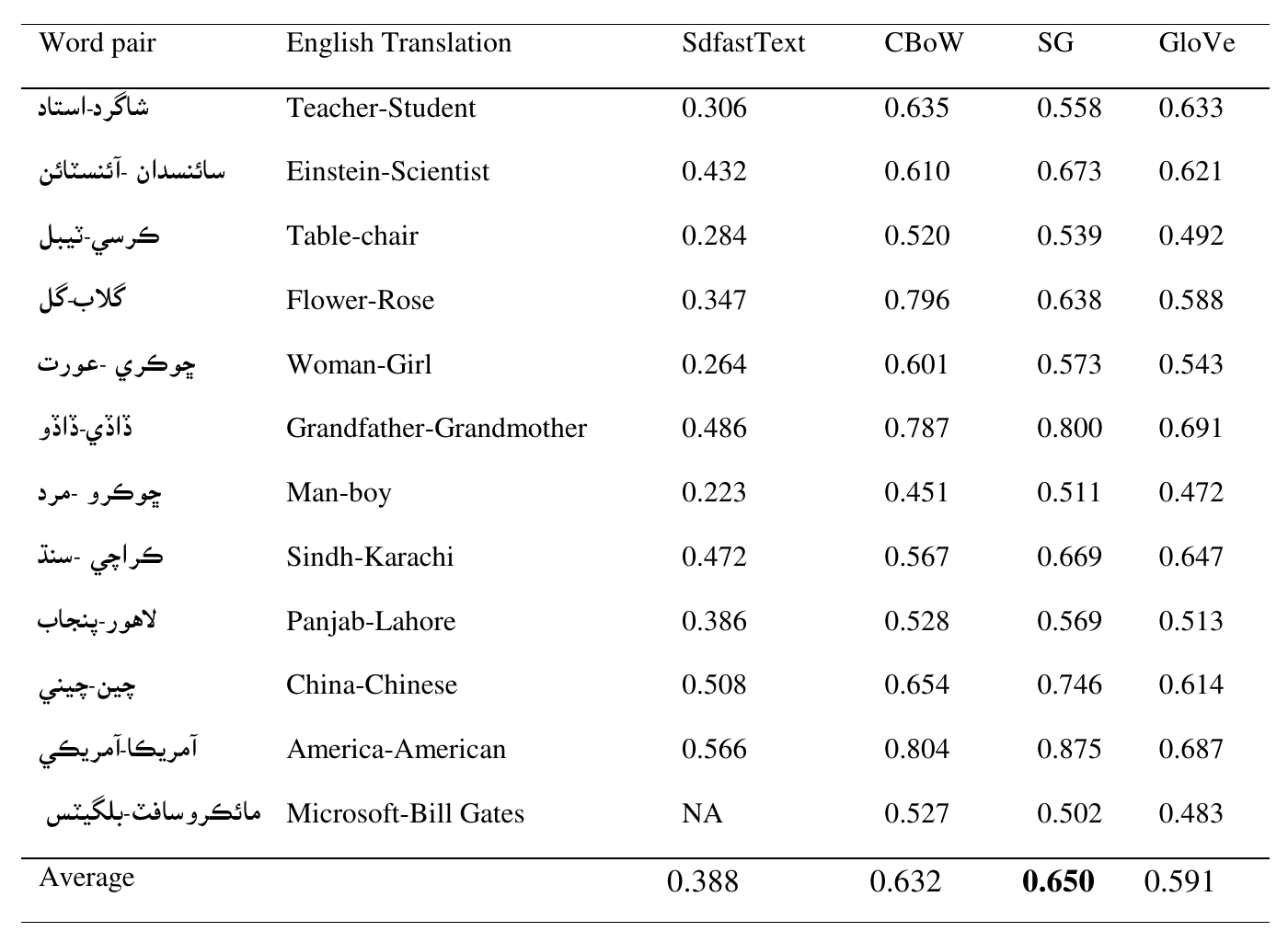}
	\caption{Word pair relationship using cosine similarity (higher is better) }
	\label{tab:word_pair_relationship}
\end{table}
\begin{table}[hthb]
	\centering
	\includegraphics[width=12cm]{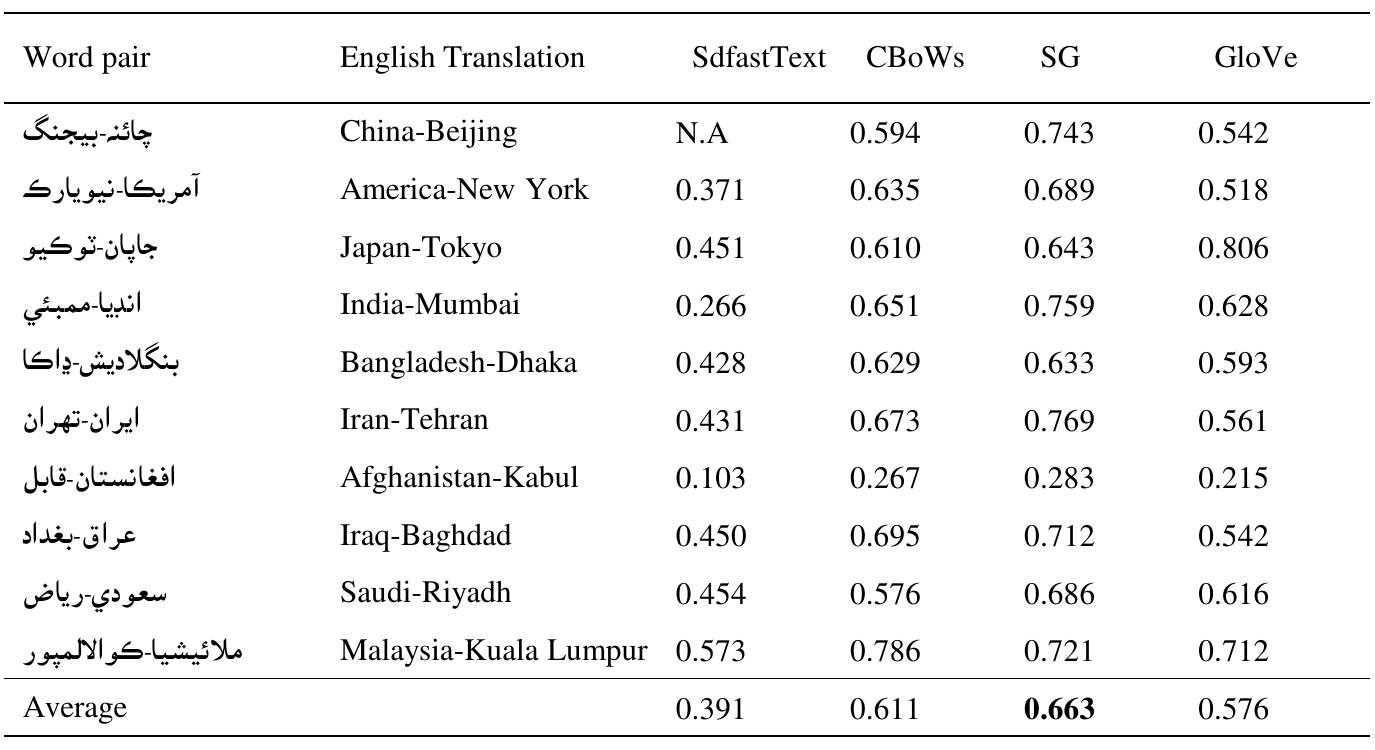}
	\caption{Cosine similarity score between country and capital}
	\label{tab:country_capital}
\end{table}
\subsection{Comparison with WordSim353 }
\label{sec:comparision_wordsim353}
We evaluate the performance of our proposed word embeddings using the WordSim353 dataset by translation English word pairs to Sindhi. Due to vocabulary differences between English and Sindhi, we were unable to find the authentic meaning of six terms, so we left these terms untranslated. So our final Sindhi WordSim353 consists of 347 word pairs. Table \ref{tab:semantic_similarity} shows the Spearman correlation results using Eq.~\ref{Spearman_corr} on different dimensional embeddings on the translated WordSim353. The Table \ref{tab:semantic_similarity} presents complete results with the different \textit{ws} for CBoW, SG and GloVe in which the \textit{ws=7} subsequently yield better performance than \textit{ws} of 3 and 5, respectively. The SG model outperforms CBoW and GloVe in semantic and syntactic similarity by achieving the performance of \textbf{0.629} with ws=7. In comparison with English \cite{mikolov2013efficient} achieved the average semantic and syntactic similarity of 0.637, 0.656 with CBoW and SG, respectively. Therefore, despite the challenges in translation from English to Sindhi, our proposed Sindhi word embeddings have efficiently captured the semantic and syntactic relationship. 
\begin{table}[htp]
    \centering
    \begin{tabular}{llll}
        \hline
        Model & $ws$ & Accuracy \\ \hline
        \multirow{4}{*}{CBoW} & 3  & 0.568 \\
        & 5  & 0.582 \\
     & 7  & 0.596 \\ \hline
     \multirow{3}{*}{Skip gram} & 3 & 0.617 \\
     & 5 &  0.621 \\
      & 7  & 0.629 \\ \hline
        \multirow{3}{*}{GloVe} & 3 & 0.542 \\
        & 5 &  0.563 \\
        & 7  & 0.568 \\ \hline
        SdfastText &  &  0.374 \\ \hline
    \end{tabular}
    \caption{Comparison of semantic and syntactic accuracy of proposed word embeddings using WordSim-353 dataset on $300-D$ embedding choosing various window size (ws).}
    \label{tab:semantic_similarity}
\end{table}
\subsection{Visualization}
\label{sec:visualization}
\begin{figure}
	\centering
	\includegraphics[width=11cm]{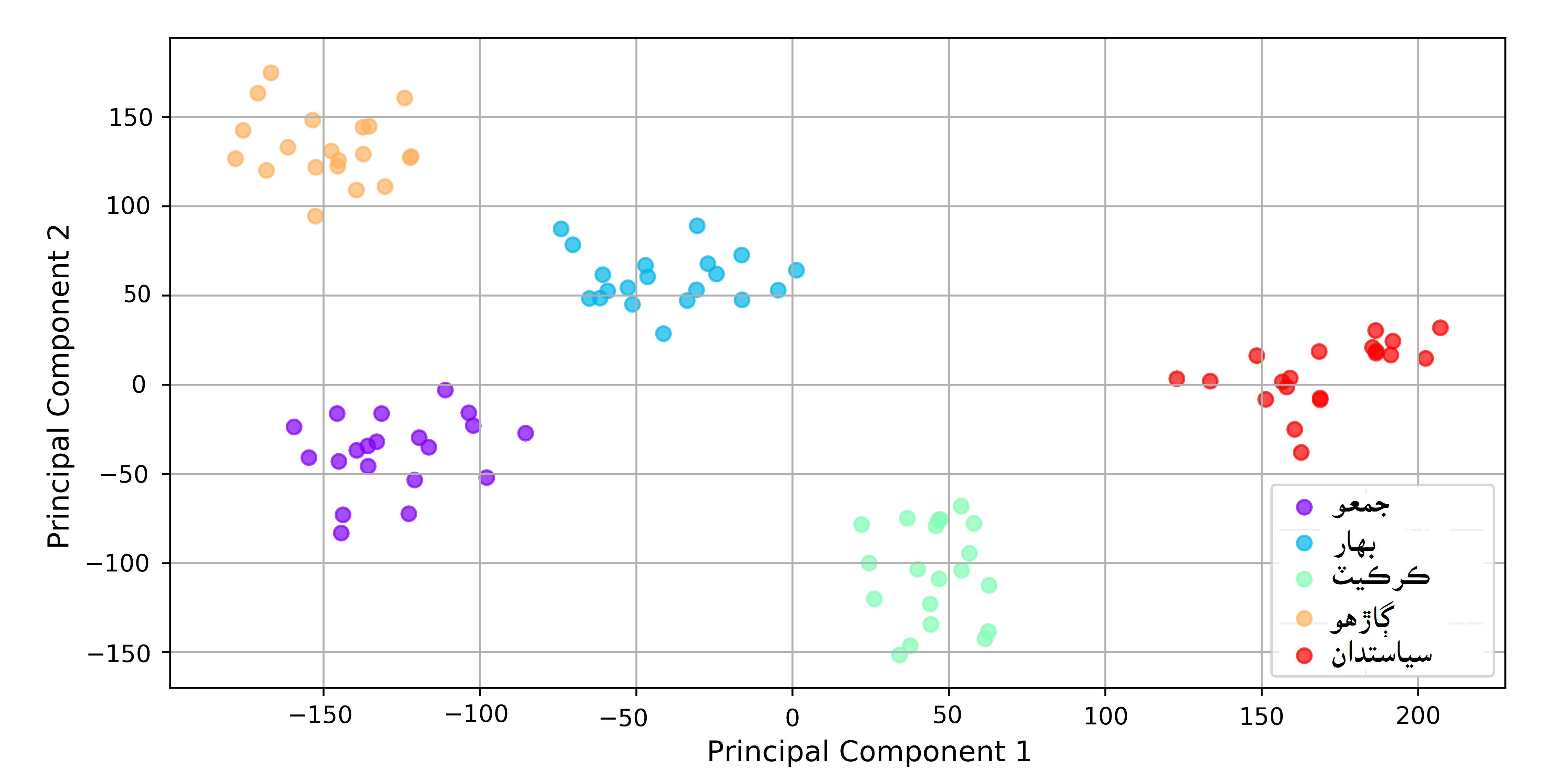}
	\caption{Visualization of Sindhi CBoW word embeddings}
	\label{fig:viz_cbow}
\end{figure}

\begin{figure}
	\centering
	\includegraphics[width=11cm]{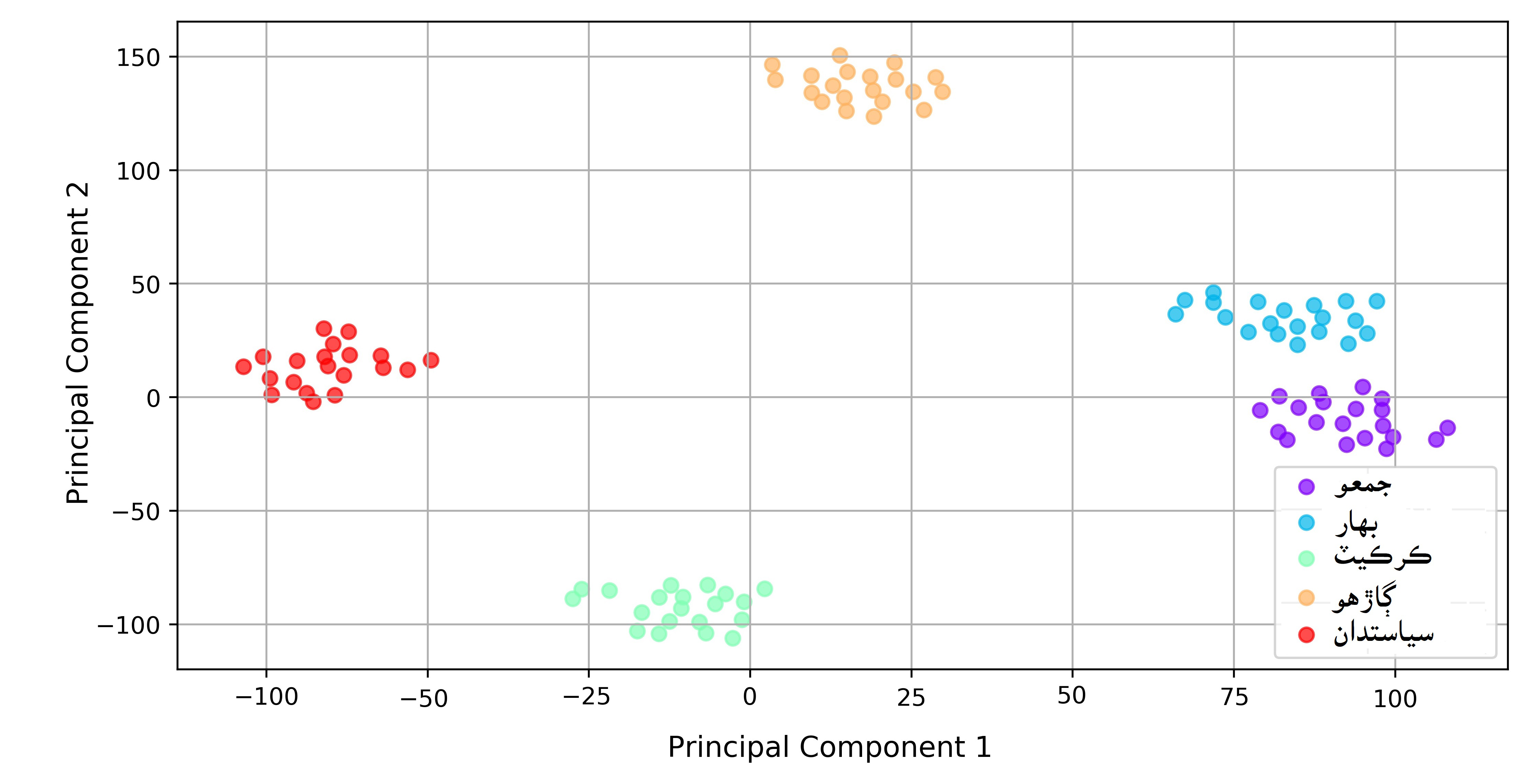}
	\caption{Visualization of Sindhi SG word embeddings}
	\label{fig:viz_SG}
\end{figure}

\begin{figure}
	\centering
	\includegraphics[width=11cm]{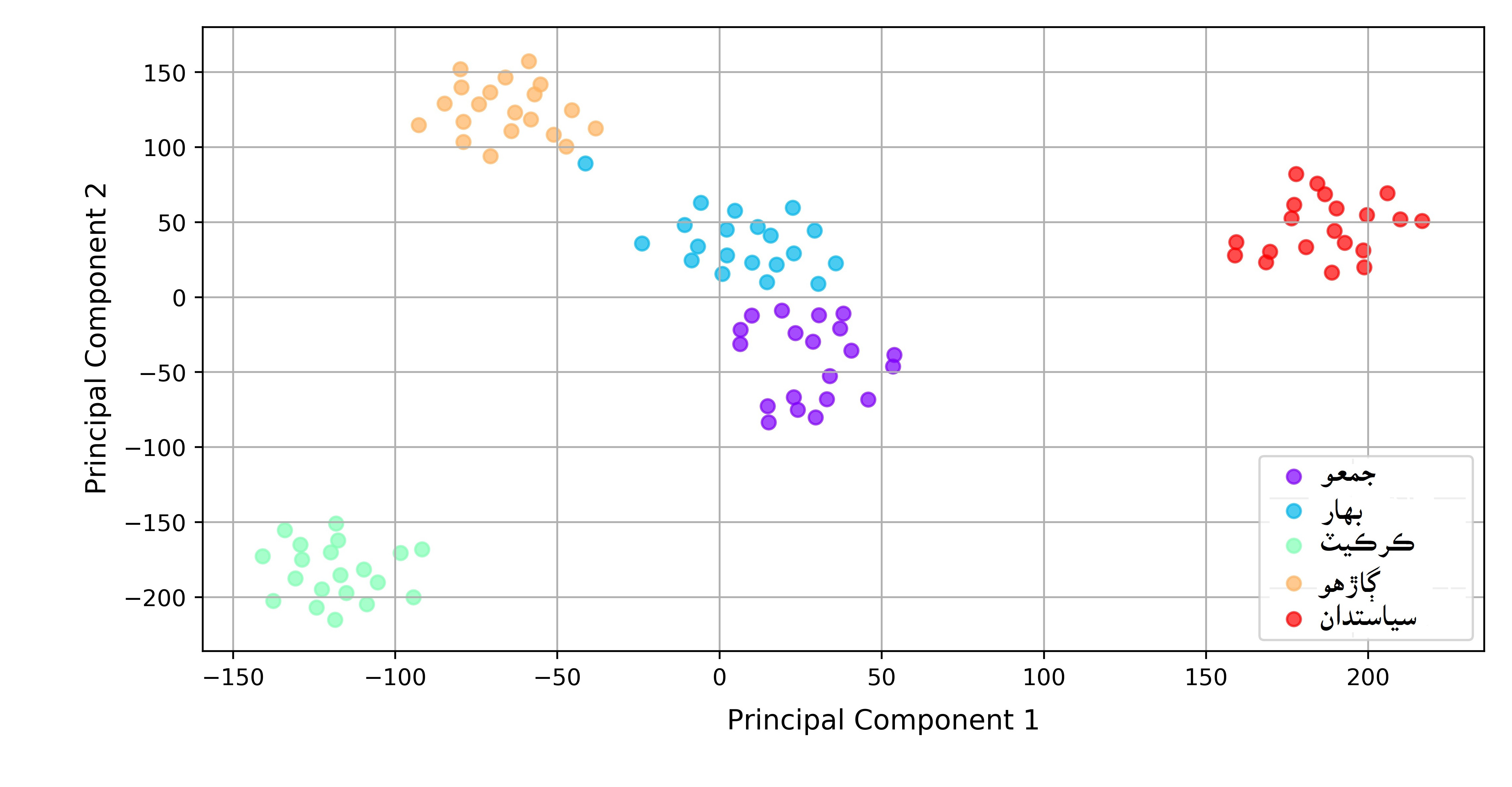}
	\caption{visualization of Sindhi GloVe word embeddings}
	\label{fig:viz_glove}
\end{figure}

\begin{figure}
	\centering
	\includegraphics[width=11cm]{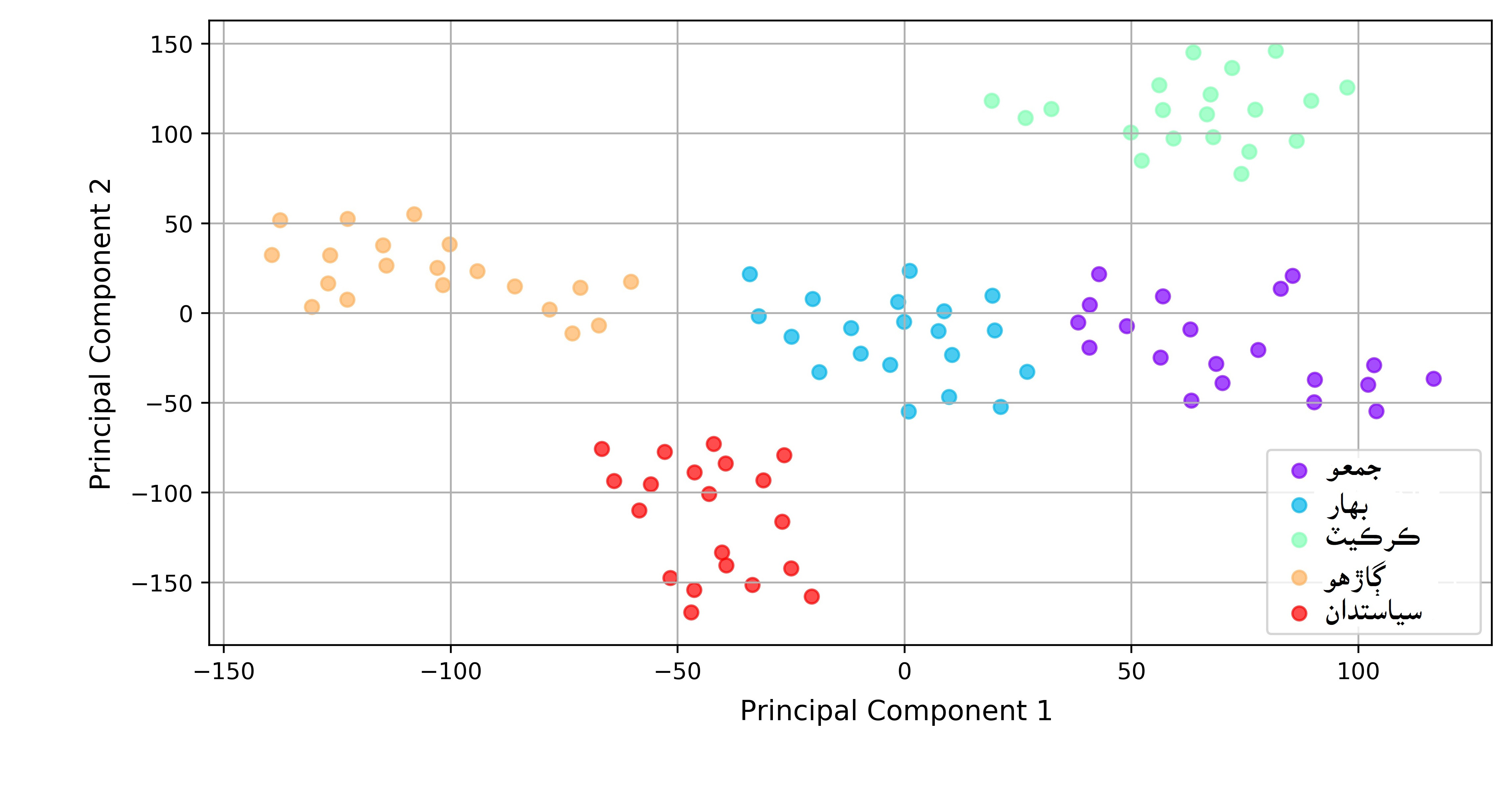}
	\caption{Visualization of SdfastText word embeddings}
	\label{fig:viz_fasttext}
\end{figure}

We use t-Distributed Stochastic Neighboring (t-SNE) dimensionality~\cite{maaten2008visualizing} reduction algorithm with PCA~\cite{lebret2013word} for exploratory embeddings analysis in 2-dimensional map. The t-SNE is a non-linear dimensionality reduction algorithm for visualization of high dimensional datasets. It starts the probability calculation of similar word clusters in high-dimensional space and calculates the probability of similar points in the corresponding low-dimensional space. The purpose of t-SNE for visualization of word embeddings is to keep similar words close together in 2-dimensional $x,y$ coordinate pairs while maximizing the distance between dissimilar words.  The t-SNE has a perplexity (PPL) tunable parameter used to balance the data points at both the local and global levels. We visualize the embeddings using PPL=20 on 5000-iterations of 300-D models. We use the same query words (see Table~\ref{tab:nearest_neighnors}) by retrieving the top 20 nearest neighboring word clusters for a better understanding of the distance between similar words. Every query word has a distinct color for the clear visualization of a similar group of words. The closer word clusters show the high similarity between the query and retrieved word clusters. The word clusters in SG (see Fig. \ref{fig:viz_SG}) are closer to their group of semantically related words. Secondly, the CBoW model depicted in Fig. \ref{fig:viz_cbow} and GloVe Fig. \ref{fig:viz_glove} also show the better cluster formation of words than SdfastText Fig. \ref{fig:viz_fasttext}, respectively.  
\section{Discussion and future work}
\label{sec:discussion}
In this era of the information age, the existence of LRs plays a vital role in the digital survival of natural languages because the NLP tools are used to process a flow of un-structured data from disparate sources. It is imperative to mention that presently, Sindhi Persian-Arabic is frequently used in online communication, newspapers, public institutions in Pakistan and India. Due to the growing use of Sindhi on web platforms, the need for its LRs is also increasing for the development of language technology tools. But little work has been carried out for the development of resources which is not sufficient to design a language independent or machine learning algorithms. The present work is a first comprehensive initiative on resource development along with their evaluation for statistical Sindhi language processing. 
\par \setlength{\parindent}{3ex} 
More recently, the NN based approaches have produced a state-of-the-art performance in NLP by exploiting unsupervised word embeddings learned from the large unlabelled corpus. Such word embeddings have also motivated the work on low-resourced languages. Our work mainly consists of novel contributions of resource development along with comprehensive evaluation for the utilization of NN based approaches in SNLP applications. The large corpus obtained from multiple web resources is utilized for the training of word embeddings using SG, CBoW and Glove models. The intrinsic evaluation along with comparative results demonstrates that the proposed Sindhi word embeddings have accurately captured the semantic information as compare to recently revealed SdfastText word vectors. The SG yield best results in nearest neighbors, word pair relationship and semantic similarity. The performance of CBoW is also close to SG in all the evaluation matrices. The GloVe also yields better word representations; however SG and CBoW models surpass the GloVe model in all evaluation matrices.
\par \setlength{\parindent}{3ex} 
Hyperparameter optimization is as important as designing a new algorithm. The choice of optimal parameters is a key aspect of performance gain in learning robust word embeddings. Moreover, We analysed that the size of the corpus and careful preprocessing steps have a large impact on the quality of word embeddings. However, in algorithmic perspective, the character-level learning approach in SG and CBoW improves the quality of representation learning, and overall window size, learning rate, number of epochs are the core parameters that largely influence the performance of word embeddings models. Ultimately, the new corpus of low-resourced Sindhi language, list of stop words and pretrained word embeddings along with empirical evaluation, will be a good supplement for future research in SSLP applications.
\par \setlength{\parindent}{3ex} 
In the future, we aim to use the corpus for annotation projects such as parts-of-speech tagging, named entity recognition. The proposed word embeddings will be refined further by creating custom benchmarks and the extrinsic evaluation approach will be employed for the performance analysis of proposed word embeddings. Moreover, we will also utilize the corpus using Bi-directional Encoder Representation Transformer~\cite{devlin2019bert} for learning deep contextualized Sindhi word representations. Furthermore, the generated word embeddings will be utilized for the automatic construction of Sindhi WordNet.
\section{Conclusion}
\label{sec:conclusion}
In this paper, we mainly present three novel contributions of large corpus development contains large vocabulary of more than 61 million tokens, 908,456 unique words. Secondly, the list of Sindhi stop words is constructed by finding their high frequency and least importance with the help of Sindhi linguistic expert. Thirdly, the unsupervised Sindhi word embeddings are generated using state-of-the-art CBoW, SG and GloVe algorithms and evaluated using popular intrinsic evaluation approaches of cosine similarity matrix and WordSim353 for the first time in Sindhi language processing. We translate English WordSim353 using the English-Sindhi bilingual dictionary, which will also be a good resource for the evaluation of Sindhi word embeddings. Moreover, the proposed word embeddings are also compared with recently revealed SdfastText word representations. 
\par \setlength{\parindent}{3ex} 
Our empirical results demonstrate that our proposed Sindhi word embeddings have captured high semantic relatedness in nearest neighboring words, word pair relationship, country, and capital and WordSim353. The SG yields the best performance than CBoW and GloVe models subsequently. However, the performance of GloVe is low on the same vocabulary because of character-level learning of word representations and sub-sampling approaches in SG and CBoW. Our proposed Sindhi word embeddings have surpassed SdfastText in the intrinsic evaluation matrix. Also, the vocabulary of SdfastText is limited because they are trained on a small Wikipedia corpus of Sindhi Persian-Arabic. We will further investigate the extrinsic performance of proposed word embeddings on the Sindhi text classification task in the future. The proposed resources along with systematic evaluation will be a sophisticated addition to the computational resources for statistical Sindhi language processing.

\bibliographystyle{unsrt}

\bibliography{reference}

\end{document}